\documentclass[letterpaper]{article}
\usepackage{aaai}
\usepackage{times}
\usepackage{helvet}
\usepackage{courier}
\usepackage{graphicx}
\usepackage{subfigure}
\usepackage{amssymb}
\usepackage{amsthm}
\usepackage{mathtools}
\usepackage{multirow}
\usepackage{tabularx}
\usepackage{makecell}
\usepackage[font=normal,skip=3pt]{caption}

\begin{document}
%
\title{Typeface Completion with Generative Adversarial Networks}
\author{Yonggyu Park,
        Junhyun Lee,
        Yookyung Koh,
        Inyeop Lee,
        Jinhyuk Lee,
        Jaewoo Kang\\
 Department of Computer Science and Engineering\\
 Korea University\\
 \{yongqyu, ljhyun33, ykko603, blackin77, jinhyuk\_lee, kangj\}@korea.ac.kr\\
}
\maketitle
\begin{abstract}
\begin{quote}
The mood of a text and the intention of the writer can be reflected in the typeface.
However, in designing a typeface, it is difficult to keep the style of various characters consistent, especially for languages with lots of morphological variations such as Chinese.
In this paper, we propose a Typeface Completion Network (TCN) which takes one character as an input, and automatically completes the entire set of characters in the same style as the input characters.
Unlike existing models proposed for image-to-image translation, TCN embeds a character image into two separate vectors representing typeface and content.
Combined with a reconstruction loss from the latent space, and with other various losses, TCN overcomes the inherent difficulty in designing a typeface.
Also, compared to previous image-to-image translation models, TCN generates high quality character images of the same typeface with a much smaller number of model parameters.

We validate our proposed model on the Chinese and English character datasets, which is paired data, and the CelebA dataset, which is unpaired data. 
In these datasets, TCN outperforms recently proposed state-of-the-art models for image-to-image translation.
The source code of our model is available at https://github.com/yongqyu/TCN.
\end{quote}
\end{abstract}
\section{Introduction}


%
%
%
%

\noindent 
Typeface is a set of one or more fonts, each consisting of glyphs that share common design features.\footnote{https://en.wikipedia.org/wiki/Typeface}. 
Effective typeface not only allows writers to better express their opinions, but also helps convey the emotions and moods of their text. 
However, there is a small number of typefaces to choose from because there are several difficulties in designing typography. The typeface of all characters should be the same without compromising readability.
As a result, it takes much effort to make a typeface for languages with a large character set such as Chinese which contains more than twenty thousand characters. 

\begin{figure}[t]
  \centering
  \includegraphics[width=\columnwidth]{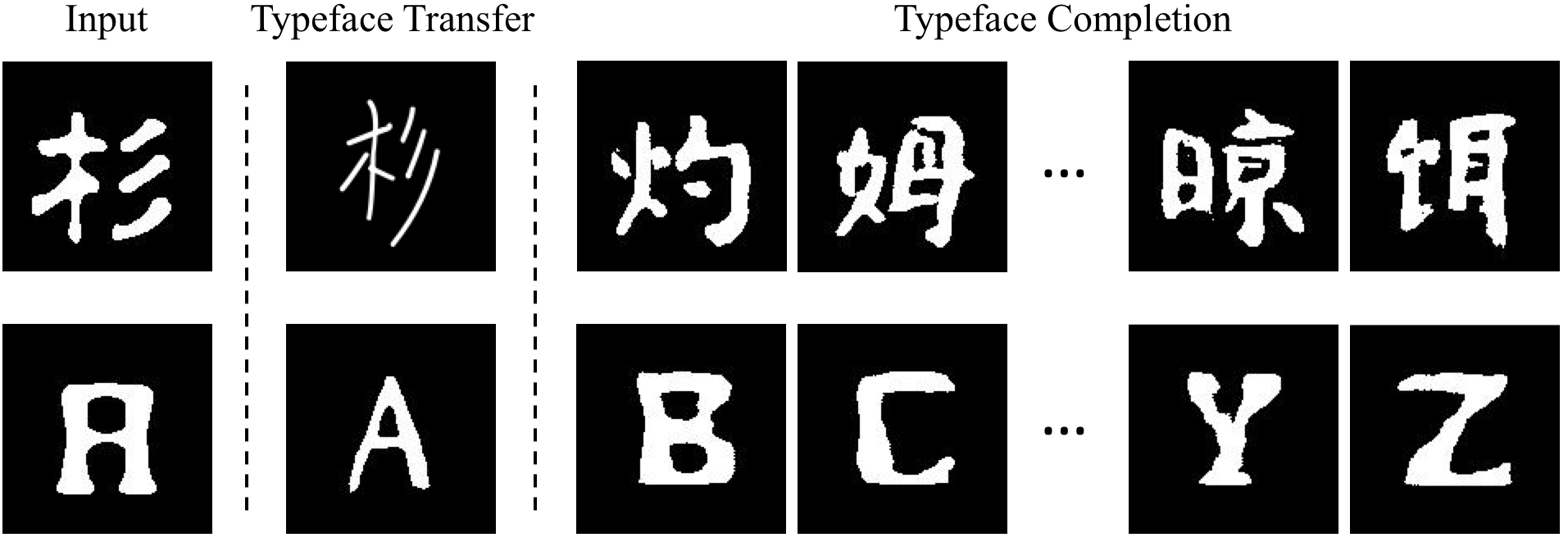}
  \caption{Comparison of the output of TCN with that of the typeface transfer model. Unlike typeface transfer model that changes the style of an input, typeface completion model generates the contents of all character sets  in the same style as the input. Typeface completion outputs are the results of TCN.}
  \label{fig:task}
\end{figure}

To deal with this difficulty, we aim to build a model that takes one character images as an input, and generates all the remaining characters in the same typeface of input characters, which is illustrated in Figure \ref{fig:task}.

In the field of computer vision, the typeface completion task has not been much studied.
Generating character images in the same typeface could be seen as a image-to-image translation problem. Existing image-to-image translation tasks often refer to extracting a style feature from a desired image, and combines the style feature while keeping the content features of desire images.
In case of typeface completion task, after extracting a style feature from a desired image, we combine the style feature while changing the content features of same desire images.
For the typeface completion task, we use the terms style feature and typeface feature, interchangeably.

And, as existing single image-to-image translation models learn only a single domain translation \cite{StyleTransfer:Gatys,StyleTransfer:Huang,StyleTransfer:Li2,StyleTransfer:Li3}, we need to train $N(N-1)/2$ models on a set of $N$ characters for the typeface completion task.
While learning all the single models and keeping them for typeface completion is computationally infeasible, recent work of Choi et al. (2017) has addressed this inefficient but fails to produce high quality character images for typeface completion.


In typeface completion, a large number of classes such as Chinese characters should be considered. Image-to-Image translation models designed for small number of classes fail to generalize in the typeface completion task due to the large number of classes in character sets.
In the existing model, the input channel becomes $1+n$ where $1$ refers to grey-scale channels for a character image, and $n$ refers to the number of classes. However, concatenating one-hot encoded labels directly to the image tensor makes the model ineffective when $n$ is large. That's because most of the input value will be zero.



In order to overcome the weakness of existing image-to-image translation models, and to deal with the large number of classes and paired dataset, we propose a Typeface Completion Network (TCN) that generates all characters in a character set from a single model.
TCN represents the typefaces and contents of characters as latent vectors, and uses various losses.
We show that TCN outperforms on Chinese and English datasets in terms of task-specific quantitative metric and image qualities. 
This is possible because the character image is paired data, unlike other general images. 
And for the same reason, we can take advantage to generate more plausible images with additional loss.
Furthermore, it also showed better performance than the existing image-to-image translation model in general images that do not take the advantage mentioned above. This shows that TCN is applicable to unpaired dataset.

\section{Related Works}
\subsection{Image-to-Image Translation}

Generative Adversarial Networks (GAN)\cite{GAN:Goodfellow} has been highlighted as one of the hottest research topics in computer vision. GAN generates images using an adversarial loss with a deep convolution architecture \cite{DCGAN:Radford,AdvanceGAN:Goodfellow}. 
GAN has gained popularity and resulted in a variety of follow-up studies \cite{CGAN:Mirza,GAN_Subtype:Perarnau,GAN_Subtype:Arjovsky,Super_resolution:Ledig,chen2017photographic} in the context of image generation, super-resolution, colorization, inpainting, attribute transfer, and image-to-image translation.


The style transfer task, one of the image-to-image translation tasks, involves changing the style of an image while retaining its content. 
Since most existing style transition models have a fixed pair of input and target style, they cannot receive or generate styles in various domains using a single model.
However, the models of \cite{CGAN:Mirza,StarGAN:Yun}, can take a target style label as an input and generate an image of the desired style using a single model. 
This reduces the number of parameters in a task which performs the transition into various style domains.

A task of completing a character set with some subsets can also be treated as a image-to-image task. 
Our model changes the content of an input to the content of the target while maintaining the style of the input. 
This is the same as the existing style transfer model where the terms, style and content, are reversed.
In addition, since our task requires various content domains, our model is based on a multi-domain transfer model.

\subsection{Character Image Generation}
Early character image generation models focused on geometric information \cite{Font:Tenenbaum,Font:Suveeranont,Font:Campbell,Font:Phan}.
But now, with the development of deep learning,
many models focus on character style transfer tasks \cite{CNNFont:Upchurch,CNNFont:Baluja}. In particular, there have been researches on the style transfer task using GAN in the character image domain \cite{GANFont:Lyu,GANFont:Chang,MCGAN:Azadi,Font2font:Bhunia}. 

However, the difference between the existing typeface transfer model with character images and our typeface completion model is that our model considers the content as a style, not a typeface, and transfer it.
This changes the number of target labels.
Due to languages with thousands of characters such as Chinese, the existing single-domain image-to-image translation model and the multi-domain image-to-image translation model which uses the one-hot vector as a domain label deal with the parameter inefficiency problem.
To solve this problem, we express the domain label as a latent vector and propose new losses accordingly.

\begin{figure*}[t]
  \centering
  \includegraphics[width=\textwidth]{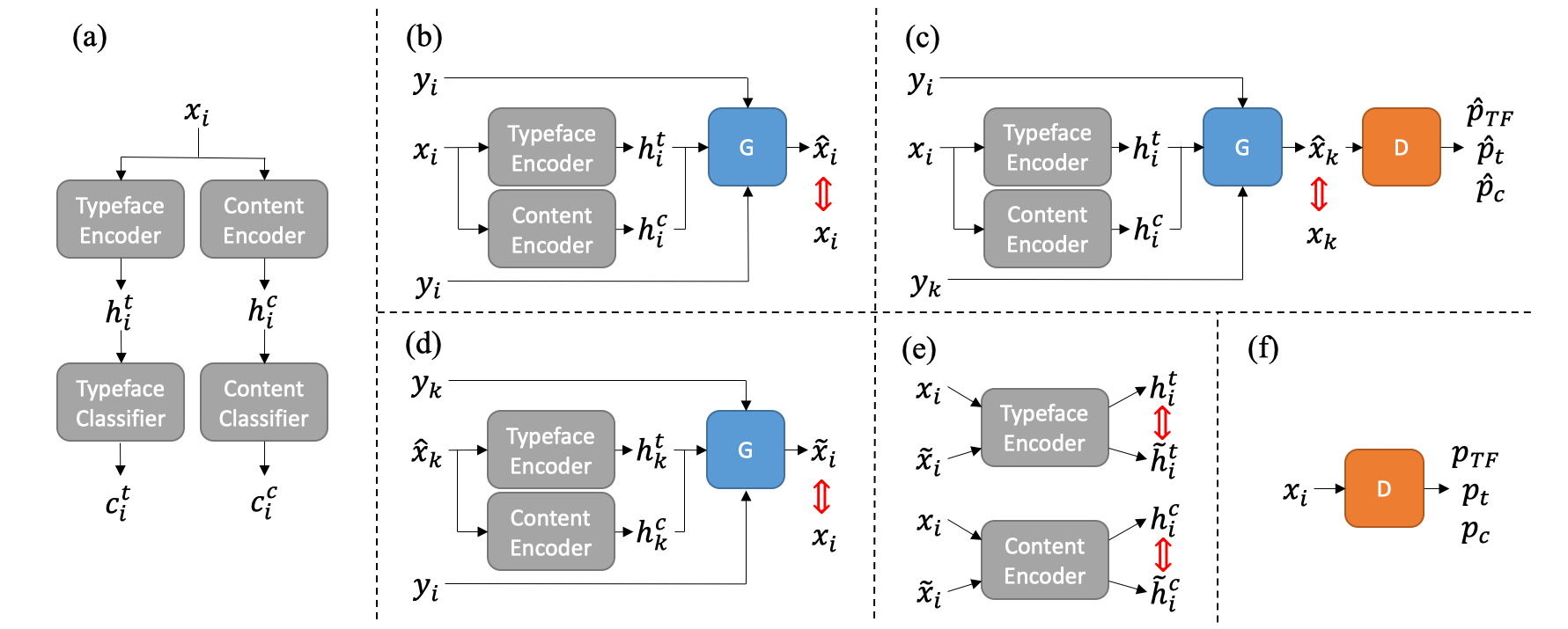}
  \caption{The overall flow chart of the TCN.
The red arrow refers to the loss from the difference. Encoders, $G$, and $D$ share the same weights in all the experiments.
(a) Encoder pretraining.
(b) Identity loss.
(c) SSIM loss comparing $\hat{x}_k$ to $x_k$, and adversarial loss leading to the result of the discriminator corresponding to $x_k$.
(d) Reconstruction loss that transforms $\hat{x}_k$ to $\tilde{x}_i$ and compares it to the original $x_i$. 
(e) Perceptual reconstruction loss.
(f) Discriminator training.
}
  \label{fig:overall}
\end{figure*}
\section{Task Definition}

The typeface completion task involves completing the remaining characters $X \setminus x_i$ of a character set $X=\{x_1, x_2, ..., x_N\}$ of a single typeface, using one of the $N$ characters, $x_i$. 
%
%
TCN receives a triplet ($x_i, y_i, y_k$) as input, where character label $y_i, y_k \in\ Y=\{y_1, y_2, ..., y_N\}$ corresponds to $x_i, x_k$. 
%
%
Using the triplet, our model generates a character image $\hat{x}_k \in\ \hat{X}=\{\hat{x_1}, \hat{x_2}, ..., \hat{x_N}\}$ , corresponding to the character of $y_k$ with the typeface of $x_i$.
%
%
%
%
Since TCN generates one character at a time, above generating process is repeated $N-1$ times. 
%
%
%
%
The goal of this task is to obtain a model parameter $\theta$ that minimizes the difference between $\hat{x}_k$ and $x_k$ while generating all the character sets. 
The overall formula of the task is as follows.

\begin{equation}\label{task1}
\theta^* = \operatorname*{argmin}_\theta \sum_{k=1, k\neq i}^N d(x_k, f(x_i, y_i, y_k, \theta))
\end{equation}
where $d$ is the distance between the images. We mainly used the $\texttt{SSIM}$ index to measure the distance between the images. From the above optimal $\theta^*$, we can obtain the $\hat{x}_k$ that is similar to $x_k$.
%
%

\begin{equation}\label{task2}
x_k \approx \hat{x}_k = f(x_i, y_i, y_k, \theta^*)
\end{equation}


\section{Proposed Model}

Figure \ref{fig:overall} shows the training process of TCN. 
%
%
TCN consists of typeface and content encoders, a discriminator, and a generator.
The encoders extract desired feature vectors from an image. 
The two encoders each return a latent vector that combines different information.
The generator, along with the two vectors above, receives character labels corresponding to the input and target. 
Through this process, the generator makes a target character image that has the same style as the input.
The discriminator receives the generated image and determines if it is real. 
In this process, the discriminator returns the probability that the image corresponds to a certain typeface and character. 


\subsection{Encoders}

In our model, the encoder is divided into a content encoder and a typeface encoder. The content encoder extracts the symbolic representation of a character from an image, and the typeface encoder extracts the typographical representation from an image. Each encoder consists of a ResNet, and a 1-layer fully connected (FC) classifier, and returns the output of the ResNet and classifier, respectively. 
%
%

\subsubsection*{Typeface and Content Feature}
The encoders receive the image from the main training and return the latent vectors containing the typeface and the content feature, respectively. Each expression is as follows:

\begin{equation}\label{e}
\begin{split}
h_i^t = E_h^t(x_i)\\
h_i^c = E_h^c(x_i).
\end{split}
\end{equation}
where the subscript $h$ of $E_h^t$ and $E_h^c$ represent the ResNets of the encoders, so each $h_i$ is a encoded latent vector of $x_i$.
%
%

\subsubsection*{Encoder Pretraining}
Since the two encoders have the same structure, we don't know what information each encoder extracts unless we guide them.
%
%
Therefore, we pretrain the encoders so that each encoder extracts disjoint information.
%
%

For pretraining, we perform the classification task to distinguish the typeface and character of an image.
%
%
The output of the classifier creates a cross entropy($\texttt{CE}$) loss, encouraging the output of each ResNet to contain corresponding feature. As shown in Figure \ref{fig:overall} (a), the losses from the typeface ($L_{\texttt{cls}}^{\texttt{te}}$) and content ($L_{cls}^{ce}$) encoders are defined as
%
%

\begin{equation}\label{e_cls}
\begin{split}
L_{\texttt{cls}}^{\texttt{te}} = \texttt{CE}(c_i^t, l_i^t), c_i^t = E_c^t(h_i^t)\\
L_{\texttt{cls}}^{\texttt{ce}} = \texttt{CE}(c_i^c, l_i^c), c_i^c = E_c^c(h_i^c)
\end{split}
\end{equation}
where the subscript $c$ of $E_c^t$ and $E_c^c$ represent the classifier of encoders, therefore each $c_i^t$ and $c_i^c$ is a classification result of the typeface and content of $x_i$, and $l_i^t$ and $l_i^c$ are the typeface and content labels of $x_i$, respectively. Especially, $l_i^c$ is equal to $y_i$.

Although the classification accuracy of each encoder is higher than 80\%, redundancy may exist between the two features to some extent because both typeface and content features are generated from the same character image 
%
%
To solve this problem, we applied triplet loss\cite{schroff2015facenet} as follows. 
\begin{equation}\label{e_triplet}
\begin{split}
L_{\texttt{triplet}}^{\texttt{te}} = ||h_i^t - h_j^t|| - ||h_i^t - h_k^t||\\
L_{\texttt{triplet}}^{\texttt{ce}} = ||h_i^c - h_k^c|| - ||h_i^c - h_j^c||
\end{split}
\end{equation}
where $h_i^t$ and $h_i^c$ are ResNet results of the typeface and content, respectively. the typeface $x_j$ is the same, the content is different from $x_i$. $x_k$ has same content as $x_i$ and has different typeface.
%
%


Last, we have obtained a reconstruction loss as in auto-encoder. The reconstruction loss ensures that no feature is lost during exclusive extraction. The $\texttt{Decoder}$ used in the reconstruction task has the same structure as our generator.
%
%
%


\subsection{Generator}
In addition to the outputs of the encoder, the generator receives input and target character labels. The generator consists of two submodules: the feature combination submodule that combines the four inputs, and the image generation submodule that generates the image using the combined inputs.
%
%
%
%

\subsubsection*{Feature Combination}
Before generating an image, we combine four inputs.
%
%
The input of the generator includes the typeface/content feature vectors of the input image ($h_i^t, h_i^c$), the character label of the input ($y_i$), and the target ($y_k$).
Ideally, the typeface encoder should extract only the typeface feature, but due to the structural nature of CNN, the typeface encoder also extracts the content feature. Accordingly, even if we extract the same typeface information from other content, we will obtain a different result.
By the combination of the inputs, we want to make the feature vectors the same as the feature vectors obtained from the target image. 
%
%
We thus made the typeface transfer task into an auto-encoder task.

Input character labels are inserted for the multi-domain task. In the multi-domain image-to-image task with various domains of input, it is helpful for the model to know the input label with the target label rather than just the target label.
%
%


%
%

We define the combination function $f$ by the following equation:
%
%
\begin{equation}\label{g_comb}
\begin{split}
u_{i \rightarrow k} = f(h_i^t, h_i^c, y_i, y_k).
\end{split}
\end{equation}
where $f$ is a 1x1 convolutional network to establish a correlation between each channel. The 1x1 convolutional network better captures correlations than concatenating vectors and requires fewer parameters than a FC layer.

\subsubsection{Image Generation}

Next, the generator creates the image after receiving the result of the feature combination. The image generation model is composed of deconvolutional model.
We define the image generation function $g$ by the following equation:
\begin{equation}\label{g_gen}
\begin{split}
\hat{x}_k = g(u_{i \rightarrow k})
\end{split}
\end{equation}
where $g$ can be seen as a generator of a vanilla GAN that takes a latent vector and generates an image.

We concatenate the two functions of the generator and define it as $G$, and it can be expressed as follows:
\begin{equation}\label{g}
\begin{split}
\hat{x}_k = G(h_i^t, h_i^c, y_i, y_k) = g \cdot f(h_i^t, h_i^c, y_i, y_k).
\end{split}
\end{equation}
We do not distinguish between $f$ and $g$ in the future, but we only use $G$.

\subsection{Discriminator}
The discriminator takes an image and determines whether the image is a real image or a fake image generated by the generator. 
This is used as a loss so that the image created by the generator will appear real enough to fool the discriminator. 

The discriminator consists of ResNet, as in the encoders. The difference between the discriminator and the encoder is the classification part. For the encoder, there is a separate ResNet for each typeface/content classifier to distinguish the typeface and content. On the other hand, the discriminator uses one ResNet and three classifiers. One returns the T/F probability of whether the image is real, just like the discriminator of the basic GAN. The others determine which typeface and content the input has, as in \cite{CGAN:Mirza,SGAN:Odena,CGAN:Odena,GAN_Subtype:Perarnau}. 
%
%
Our discriminator did not use a separate ResNet for each classifier and thus uses fewer parameters and normalizes losses for the three tasks. 
Another difference is that our discriminator does not return the output of ResNet because it is not necessary.

\begin{equation}\label{d}
\begin{split}
p_{\texttt{TF}}, p_t, p_c = D(x_i)\\
\hat{p}_{\texttt{TF}}, \hat{p}_t, \hat{p}_c = D(\hat{x}_i)
\end{split}
\end{equation}
We define the discriminator as $D$ and express the results that correspond to the real image and fake image. The $\hat{}$ denotes to be associated with a fake image by the generator.

\subsection{Training Process}
\subsubsection{Identity Loss}
Identity loss is similar to the loss in the auto-encoder in that it helps an output to be equal to an input (Fig. \ref{fig:overall} (b)). The generator uses the character label of an input image as an input label and a target label. This experiment prevents possible loss during feature compression.
%
%

\begin{equation}\label{g_id}
L_{\texttt{id}} = ||x_i - \hat{x}_i||_1
\end{equation}
where $\hat{x_i}$ is the image generated so that has same typeface and content with $x_i$.

\subsubsection{SSIM Loss}
Structural SIMilarity index (SSIM) is used to measure the structural similarity between two images. We use SSIM index as an evaluation metric for the performance, and we also use it as a loss (Fig. \ref{fig:overall} (c) red arrow). Using SSIM index as a loss was proposed in \cite{SSIMLoss:Zhao,SSIMLoss:Snell}. In our experiments, we applied l1-loss along with the SSIM index, in the way that showed the best performance at \cite{SSIMLoss:Zhao}.

\begin{equation}\label{g_ssim}
L_{\texttt{ssim}} = ||x_k - \hat{x}_k||_1 - \texttt{SSIM}(x_k, \hat{x}_k)
\end{equation}
The \texttt{SSIM} function returns the SSIM index between two inputs. Detailed formula is in session 5.2.1.

\subsubsection{Adversarial Losses}
Adversarial losses that help outputs to look real and deceive the discriminator are the ones that are same as those of vanilla GAN. Additionally, there is also a typeface/content classification loss between the true typeface/content label and the output that the discriminator returns. We can show these losses at the end of Figure \ref{fig:overall} (c).
%
%

\begin{align}\label{g_adv}
\begin{split}
L_{\texttt{gan}} = \texttt{CE}(\hat{p}_{\texttt{TF}}, l_k^{\texttt{TF}})
\end{split}
\end{align}
\begin{align}\label{g_cls}
\begin{split}
L_{\texttt{cls}} = {}&\texttt{CE}(p_{t}, l_k^t)+\texttt{CE}(p_{c}, l_k^c)
\end{split}
\end{align}
In generator training, input image is always fake (Fig. \ref{fig:overall} (c)), but since it should deceive discriminator, $l_i^{\texttt{TF}}$ set to 1.

\subsubsection{Reconstruction Loss}
Reconstruction loss, proposed by CycleGAN\cite{Cycle_GAN:Zhu}, is a loss between the original image and the reconstruction image that is translated back to the original image from the typeface-changed image  (Fig. \ref{fig:overall} (d)). To calculate the reconstruction loss, we use the following loss function: 
\begin{align}\label{g_rec}
\begin{split}
L_{\texttt{rec}} = {}&||x_i - \tilde{x}_i||_1
\end{split}
\end{align}
where $\tilde{x}_i$ is the reconstruction image which has the same typeface and content as $x_i$.

\subsubsection{Perceptual Reconstruction Loss}
Reconstruction loss was proposed for a pixel by pixel comparison between images, but we also apply perceptual loss to this concept. A perceptual loss was first proposed by \cite{StyleTransfer:Johnson} in the style transfer field. This loss compares high-dimensional semantic information in the feature vector space. Since a character image is an image composed of strokes rather than pixel units, it is appropriate to apply the perceptual loss for reconstruction image, as shown Fig. \ref{fig:overall} (e). The equation is as follows:

\begin{equation}\label{g_per}
L_{\texttt{per}} = ||h_i^t - \tilde{h}_i^t||_2^2 
		+ ||h_i^c - \tilde{h}_i^c||_2^2
\end{equation}
where $\tilde{h}_i^t$ and $\tilde{h}_i^c$ is the output of the $E_z^t$ and $E_z^c$ for $\hat{x}_i$.

Perceptual reconstruction loss is the difference between the outputs of the encoder. We compare the input image and the image translated twice, not once. The typeface of the input image and that of the image translated once are the same. However, applying perceptual loss to the two images is not effective because these two images have different content features. Hence, we compare the input image and the twice-translated image with the same typeface/content as the input image.
%
%
%
%


The final loss of the generator is as follows:

\begin{align}
  \phantom{i + j + k}
  &\begin{aligned}
    \mathllap{L_g} &=
    L_{\texttt{gan}} +
    \lambda_{\texttt{cls}} L_{\texttt{cls}}\\
    &\qquad+ \lambda_{\texttt{ssim}} L_{\texttt{ssim}}\\
    &\qquad+ \lambda_{\texttt{rec}} (L_{\texttt{rec}} + L_{\texttt{per}} + L_{\texttt{id}})
  \end{aligned}
\end{align}
where $\lambda_{\texttt{cls}}$, $\lambda_{\texttt{ssim}}$ and $\lambda_{\texttt{rec}}$ are hyper-parameters that control the importance of each loss.

\subsubsection{Discriminator Loss}
The learning method of the discriminator is similar to that of the existing GAN. The discriminator receives two types of input: one is a real image and the other is a fake image generated by the generator. For real images, the model computes the classification loss using T/F, typeface, and content output. For fake images, only the classification loss of the T/F output is calculated because the discriminator does not need to take a loss for poor images that the generator makes.
%
%
%
%

\begin{align}\label{g_d}
\begin{split}
L_{d} = {}&\texttt{CE}(\hat{p}_{\texttt{TF}}, l_i^{\texttt{TF}})+\texttt{CE}(p_{\texttt{TF}}, l_i^{\texttt{TF}})\\
		  {}&+\texttt{CE}(p_{t}, l_i^t)+\texttt{CE}(p_{c}, l_i^c)
\end{split}
\end{align}
where $l_i^{\texttt{TF}}$ is 1 if the input image is real and otherwise 0.
$l_i^{\texttt{TF}}$ will be 0 at Figure \ref{fig:overall} (c) because the discriminator receives a fake image, and will be 1 at Figure \ref{fig:overall} (f) because $x_i$ is real.
 
\subsection{Test Process}
In training, as shown in Figure 2, we induced the losses through several steps, but in the test, we carry out one step, with only encoders and generator, not using discriminator. The typeface completion task in the test is expressed as follows:
%
%

\begin{align}\label{g_test}
\begin{split}
x_k \approx \hat{x_k} = G(h_i^t, h_i^c, y_i, y_k)
\end{split}
\end{align}
By repeating this equation $N-1$ times according to $k$, we can complete one typeface consisting of $N$ characters.
%
%
\section{Evaluation}
\subsection{Datasets}
\subsubsection{Chinese Character}
Since there are more than 50K characters in Chinese, we chose the top 1,000 most used characters\footnote{http://www.qqxiuzi.cn/zh/xiandaihanyu-changyongzi.php}. 
Chinese images were collected from true-type format (TTF) and open-type format (OTF) files obtained from the Web\footnote{https://chinesefontdesign.com}\footnote{http://www.sozi.cn}. A total of 150 files were manually selected.
%
%
Since all files do not contain all of the 1,000 characters, we have a dataset with a total of 137,839 character images. All character image sizes are 128x128, and are gray-scale 1-channel images.
%
%

\subsubsection{English Character}
We also build an English dataset for comparison. 
We used a total of 907 typographies and 26 uppercase characters. As a result of using the same selection process as the Chinese dataset, we obtained 23,583 images in total. The detailed composition  is shown in Table 1. 

\subsubsection{CelebA}
We performed a style transition experiment on the CelebA\cite{liu2015deep} dataset to measure the performance of TCN. We used 202,599 images and resized them all to 128x128, as was done with the other dataset. We used three features: black, blond, brown hair colors. The data composition and other settings are the same as those of the baseline \cite{StarGAN:Yun}.
%
%
%
%

\begin{table}[t]
\small
\begin{center}
\caption{Dataset composition}
\begin{tabular}[width=\columnwidth]{c|c|c|c|c}
\hline
\textbf{} & \multicolumn{2}{c|}{Chinese} & \multicolumn{2}{c}{English}\\\cline{2-5}
\textbf{} & \#Typeface& \#Image & \#Typeface & \#Image\\ \hline
$Train$ & 105 & 96,426 & 635 & 16,495\\ \hline
$Validation$ & 15 & 13,776 & 90 & 2,357\\ \hline
$Test$ & 30 & 27,637 & 181 & 4,733\\ \hline
\end{tabular}
\end{center}
\label{Dataset composition}
\end{table}


\subsection{Metrics}
\subsubsection{SSIM}
Unlike general images, a dataset of character images can be used to evaluate the output using an objective metric because character data has all the input-target pairs. 
%
%
We used the Structural Similarity (SSIM) index to objectively evaluate the  performance on the character data. SSIM is a metric that measures the quality of images using structural information, and is defined by the following equation:
%
%
\begin{equation}
\texttt{SSIM}(a,b)= \frac{(2\mu_a\mu_b+c_1)(2\sigma_{ab}+c_2)}{(\mu_a^2+\mu_b^2+c_1)(\sigma_a^2+\sigma_b^2+c_2)}
\end{equation}
where $\mu_a$ is the average value of the $a$ which denotes the brightness of the image. $\sigma_a$ is the distribution of the $a$ which denotes the contrast ratio of the image. $\sigma_{ab}$ is the covariance of $a$ and $b$, which denotes the correlation of the two images. $c_1$ and $c_2$ are small constants that prevent the denominator from being zero.
%
%
%
%
The closer the score is to 1, the more similar the image is to the original image.

\subsubsection{L1 distance}
In general, the $L1$ distance is used in computer vision area \cite{8195348}. The $L1$ distance is pixel-wise difference between the generated image and the target image. We can measure the pixel-wise performance with a intuitive and easily implemented way. 

\begin{table}[t]
\small
\begin{center}
\caption{Accuracy of classifier}
\label{Accruacy of classifier}
\begin{tabular}[width=\columnwidth]{c|c|c}
\hline
 & Typeface & Content \\ \hline
$Chinese$ & 99.4 & 99.7 \\ \hline
$English$ & 100.0 & 100.0 \\ \hline
\end{tabular}
\end{center}
\end{table}

\subsubsection{Classification Accuracy}
We introduce the classification accuracy as metric used in  \cite{chang2018generating}. At test dataset, We train the typeface and the content classifier, which are the ResNet. Because these classifiers show high performance accuracy over each dataset, it is reliable to be used for the metric. Each accuracy of each dataset can see at Table \ref{Accruacy of classifier}.

\subsection{Implementation Details}
We selected the learning strategy and  hyper-parameters of the models for the experiment. 
 

The encoder, discriminator, and generator are all trained using the Adam optimizer, with a learning rate of 0.0001, beta1 = 0.5, beta2 = 0.999. The learning rate gradually decreases to zero as the number of epochs is increased. 
The dimensions of $h_i$ and $u_{i \rightarrow k} \in \mathbb{R}^{256}$.
$\lambda_{\texttt{cls}}$, $\lambda_{\texttt{ssim}}$ is 5, $\lambda_{\texttt{rec}}$ is 10.
In fact, $\lambda_{texttt{cls}}$ can be assigned differently for typeface and content.
This ratio is a coefficient of trade-off between the typeface accuracy and the content accuracy of the resulting image.
This ratio acts as a trad-off coefficient between typeface accuracy and content accuracy.
The source code, implemented with Pytorch\cite{Pytorch}, is also available at https://github.com/yongqyu/TCN.

\subsection{Baselines}

\begin{table}[t]
\small
\begin{center}
\caption{Baselines}
\begin{tabular}[width=\columnwidth]{c|c|c}\hline
 & \makecell{Multi-Domain \\ Available} & \makecell{Rep. of \\ Domain Index} \\ \hline
CycleGAN & X & None \\ \hline
MUNIT & X & \makecell{Real-valued\\Distributed} \\ \hline
StarGAN & O & \makecell{One-hot} \\ \hline
TCN & O & \makecell{Real-valued\\Distributed} \\ \hline
\end{tabular}
\end{center}
\label{Baselines}
\end{table}

\subsubsection{CycleGAN}
In recent years, CycleGAN\cite{Cycle_GAN:Zhu} has obtained outstanding performance in the image-to-image translation task. CycleGAN was the first to use cycle consistency which 
makes the image that was once converted back to the original domain equal to the original image, as in Equation \ref{g_rec}. 
We also use vector-wise cycle consistency and pixel-wise cycle consistency at the image level.

\subsubsection{MUNIT}
MUNIT\cite{MUNIT:Huang} extracts the typeface and content features as a form of a latent vector using each encoder. 
%
%
By switching these vectors,  an image with the desired features can be obtained.
MUNIT uses a latent vector, and the reconstruction loss that uses the latent vector is similar to our perceptual reconstruction loss. However, there is a difference in the concept of reconstruction: we translate twice so that the reconstructed image is the same as the original, but MUNIT translates only once.
Another difference is that MUNIT needs two images to generate every image. This is not only dependent on content input image, but also inefficient if the target label is fixed.
%
%
%
%

\subsubsection{StarGAN}
StarGAN\cite{StarGAN:Yun} passes an image with the desired domain label to the generator, like cGAN. To this end, the discriminator returns the true likelihood of the image, along with the domain to which the image corresponds. As a result, StarGAN can generate all characters in one model, like our model. However, unlike our model, StarGAN uses a one-hot vector to represent a content vector. 
The comparison of the above baselines and TCN is summarized in Table 3.

%
%
%

\subsection{Experiment}
The single-domain transfer models cannot generate the entire character set using one model.
For a fair comparison with the single-domain transfer model, we used two experimental conditions. First, we used sample pairs of characters. In English, Y-G and Q-G pairs were selected to represent the most different and similar pairs, respectively.
%
%
In Chinese, index number 598-268, and 598-370 pairs were selected.
%
%
%
%
After the sample pair experiments, we compared the performance of our model with that of a multi-domain model on translating all pairs. 
In these two experimental conditions, we performed the following subtasks: Typeface Completion and Character Reconstruction.

\subsubsection*{Typeface Completion}


In the image-to-image task, our model takes one character image and learns to complete the rest of the character set while maintaining its typeface. We used a single character image as an input for a fair comparison with the other models.
%
%
We trained our model on Chinese and English character sets, which we mentioned above. 
%
%

The image-to-image translation experiment allows us to evaluate the performance of the two encoders in extracting the disjoint features. If the typeface encoder extracts the content feature and the typeface feature, the generated image will have the same content of the typeface input. This also applies to content encoders. In training, the model processes every content of a character set. And in the test, every content of the character set can be generated using the extracted typeface feature, even if the input typeface is new to the typeface encoder. Since the character image set has the target pair and the input, we can objectively evaluate the result based on its score.
%
%
%
%
%
%
%
%
%
%
%
%

\subsubsection*{Character Reconstruction}
Reconstruction is the process of regenerating an input image using the typeface and content features extracted from the input image. By the reconstruction, we can check if there is any feature missing when the encoder extracts features. It is also possible to check whether the decoder can effectively combine the two types of features. However, in this task, it is not possible to verify whether each of the features is disjointed or overlapped.

\begin{table*}[t]
\small
\begin{center}
\caption{Ablation Study Results}
\begin{tabular}[width=\columnwidth]{c|c||c|c|c|c|c|c|c|c}
\hline
Input & Target& TCN& (-)$L_{ssim}$& (-)$L_{id}$& (-)$L_{ssim} + L_{id}$& (-)$L_{rec}$& (-)$L_{per}$& (-)$L_{rec} + L_{per}$& (-)$y_{input}$  \\ \hline
\begin{minipage}{10mm}
      \includegraphics[width=10mm, height=10mm]{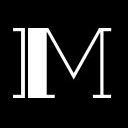}
    \end{minipage} & 
\begin{minipage}{10mm}
      \includegraphics[width=10mm, height=10mm]{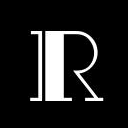}
    \end{minipage} & 
\begin{minipage}{10mm}
      \includegraphics[width=10mm, height=10mm]{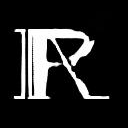}
    \end{minipage} & 
\begin{minipage}{10mm}
      \includegraphics[width=10mm, height=10mm]{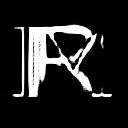}
    \end{minipage} & 
\begin{minipage}{10mm}
      \includegraphics[width=10mm, height=10mm]{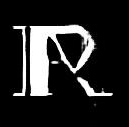}
    \end{minipage} & 
\begin{minipage}{10mm}
      \includegraphics[width=10mm, height=10mm]{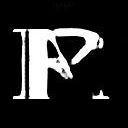}
    \end{minipage} &
\begin{minipage}{10mm}
      \includegraphics[width=10mm, height=10mm]{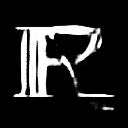}
    \end{minipage} & 
\begin{minipage}{10mm}
      \includegraphics[width=10mm, height=10mm]{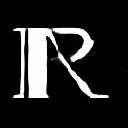}
    \end{minipage} &
\begin{minipage}{10mm}
      \includegraphics[width=10mm, height=10mm]{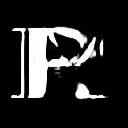} 
      \end{minipage}&
\begin{minipage}{10mm}
      \includegraphics[width=10mm, height=10mm]{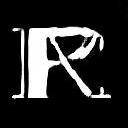}
    \end{minipage} \\\hline
    
\begin{minipage}{10mm}
      \includegraphics[width=10mm, height=10mm]{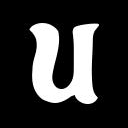}
    \end{minipage} & 
\begin{minipage}{10mm}
      \includegraphics[width=10mm, height=10mm]{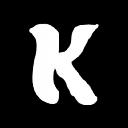}
    \end{minipage} & 
\begin{minipage}{10mm}
      \includegraphics[width=10mm, height=10mm]{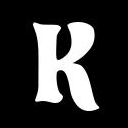}
    \end{minipage} & 
\begin{minipage}{10mm}
      \includegraphics[width=10mm, height=10mm]{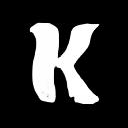}
    \end{minipage} & 
\begin{minipage}{10mm}
      \includegraphics[width=10mm, height=10mm]{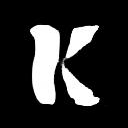}
    \end{minipage} & 
\begin{minipage}{10mm}
      \includegraphics[width=10mm, height=10mm]{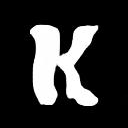}
    \end{minipage} &
\begin{minipage}{10mm}
      \includegraphics[width=10mm, height=10mm]{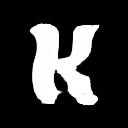}
    \end{minipage}& 
\begin{minipage}{10mm}
      \includegraphics[width=10mm, height=10mm]{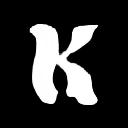}
    \end{minipage} &
\begin{minipage}{10mm}
      \includegraphics[width=10mm, height=10mm]{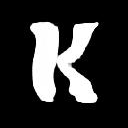} 
      \end{minipage}&
\begin{minipage}{10mm}
      \includegraphics[width=10mm, height=10mm]{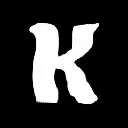}
    \end{minipage} \\\hline
    
\begin{minipage}{10mm}
      \includegraphics[width=10mm, height=10mm]{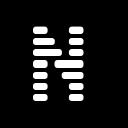}
    \end{minipage} & 
\begin{minipage}{10mm}
      \includegraphics[width=10mm, height=10mm]{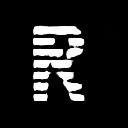}
    \end{minipage} & 
\begin{minipage}{10mm}
      \includegraphics[width=10mm, height=10mm]{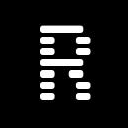}
    \end{minipage} & 
\begin{minipage}{10mm}
      \includegraphics[width=10mm, height=10mm]{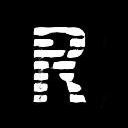}
    \end{minipage} & 
\begin{minipage}{10mm}
      \includegraphics[width=10mm, height=10mm]{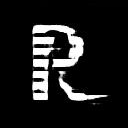}
    \end{minipage} & 
\begin{minipage}{10mm}
      \includegraphics[width=10mm, height=10mm]{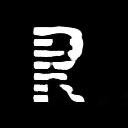}
    \end{minipage} &
\begin{minipage}{10mm}
      \includegraphics[width=10mm, height=10mm]{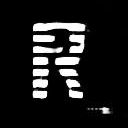}
    \end{minipage}& 
\begin{minipage}{10mm}
      \includegraphics[width=10mm, height=10mm]{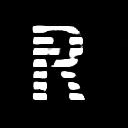}
    \end{minipage} &
\begin{minipage}{10mm}
      \includegraphics[width=10mm, height=10mm]{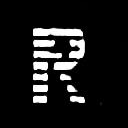} 
      \end{minipage}&
\begin{minipage}{10mm}
      \includegraphics[width=10mm, height=10mm]{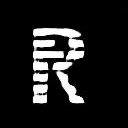}
    \end{minipage} \\\hline
    
\begin{minipage}{10mm}
      \includegraphics[width=10mm, height=10mm]{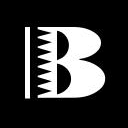}
    \end{minipage} & 
\begin{minipage}{10mm}
      \includegraphics[width=10mm, height=10mm]{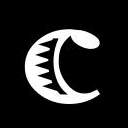}
    \end{minipage} & 
\begin{minipage}{10mm}
      \includegraphics[width=10mm, height=10mm]{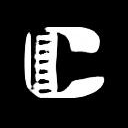}
    \end{minipage} & 
\begin{minipage}{10mm}
      \includegraphics[width=10mm, height=10mm]{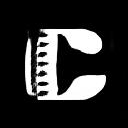}
    \end{minipage} & 
\begin{minipage}{10mm}
      \includegraphics[width=10mm, height=10mm]{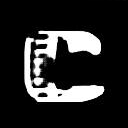}
    \end{minipage} & 
\begin{minipage}{10mm}
      \includegraphics[width=10mm, height=10mm]{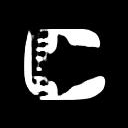}
    \end{minipage} &
\begin{minipage}{10mm}
      \includegraphics[width=10mm, height=10mm]{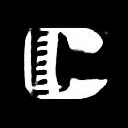}
    \end{minipage}& 
\begin{minipage}{10mm}
      \includegraphics[width=10mm, height=10mm]{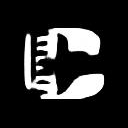}
    \end{minipage} &
\begin{minipage}{10mm}
      \includegraphics[width=10mm, height=10mm]{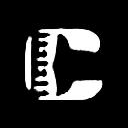} 
      \end{minipage}&
\begin{minipage}{10mm}
      \includegraphics[width=10mm, height=10mm]{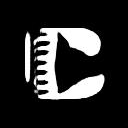}
    \end{minipage} \\\hline\hline
    
\multicolumn{2}{c||}{$SSIM$} & 0.793& 0.765& 0.784& 0.780& 0.771& 0.773& 0.781& 0.784  \\ \hline
\end{tabular}
\caption{
All columns have input as the first column, From left to right: TCN, eliminate SSIM loss, eliminate Identity loss, eliminate SSIM and Identity loss, eliminate reconstruction loss, eliminate perceptual reconstruction loss, eliminate reconstruction and perceptual reconstruction loss. And lastly, generate without target label.
 }
\end{center}
\label{Experimental_Results}
\end{table*}

\subsubsection*{Ablation Study}
We conducted an ablation study to check for redundancy among the various losses in our model. By comparing the performance of eliminating each loss and the performance of the entire model, we can check the influence of each loss. In addition to loss, we also conducted an ablation study on the sub-modules to check the influence of the sub-modules. The ablation study were conducted in Chinese and English datasets, and results were quantitatively and qualitatively evaluated.

\subsubsection*{Face Generation}
As our model is not limited to character images, we experimented with facial images used for existing image-to-image models. In the facial image experiment, the differences from the character images are that there are no content labels and no target images. Therefore we proceeded the experiment after removed the associated losses and sub-module of TCN. We take a face image and perform a image-to-image translation experiment that changes the style feature label. We also conducted weighted image-to-image translation experiments on weighted style feature labels. Since we cannot quantitatively evaluate in the unpaired dataset, we conducted a quantitative evaluation through comparison only.
%
%

\begin{figure}[t]
  \centering
  \includegraphics[width=\columnwidth]{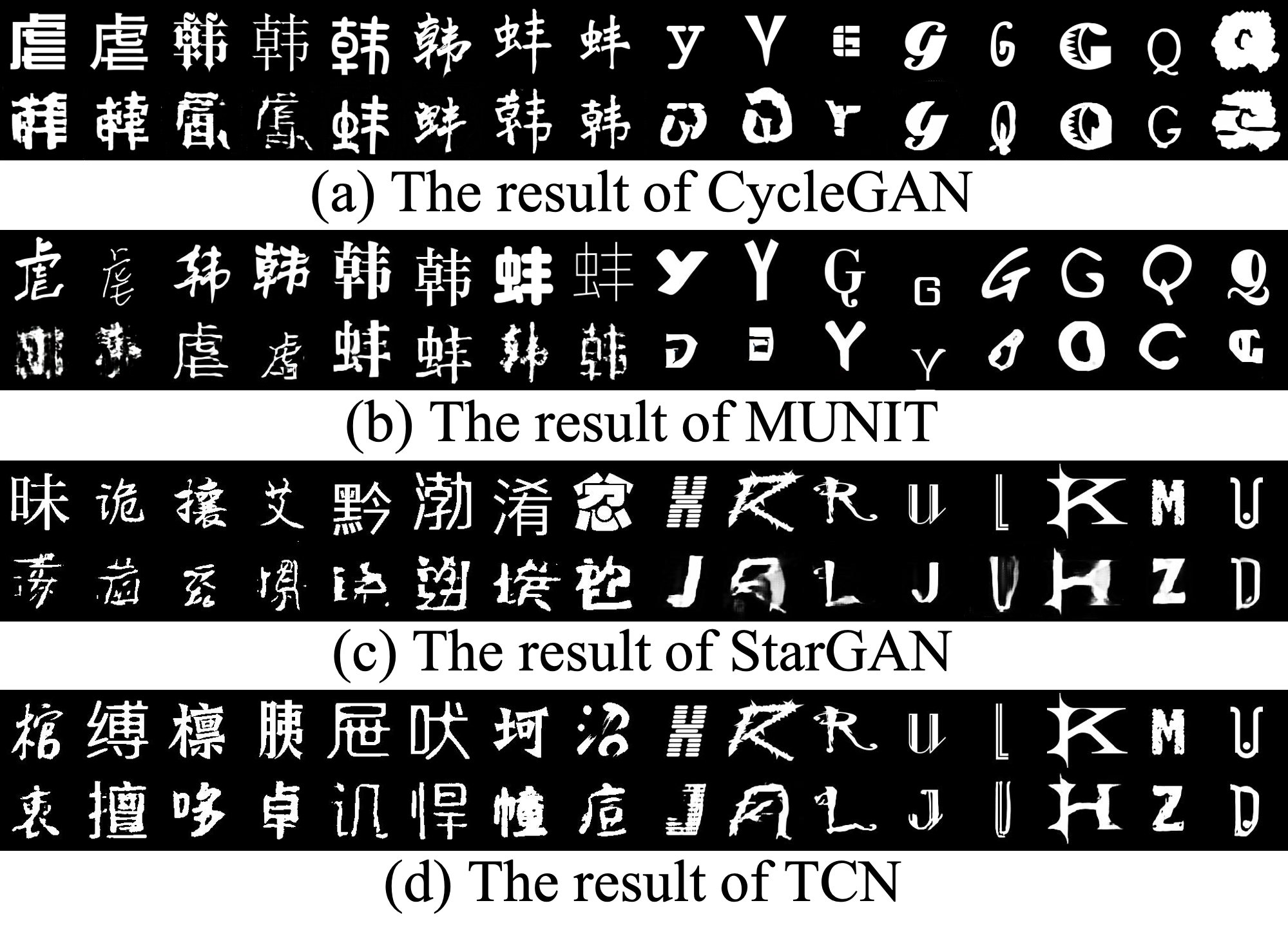}
  \caption{
 The results of baselines and our model. The first row contains the typeface source image, the second row contains the output.
%
%
 }
  \label{fig:results}
\end{figure}

\begin{figure}[ht]
  \centering
  \includegraphics[width=\columnwidth]{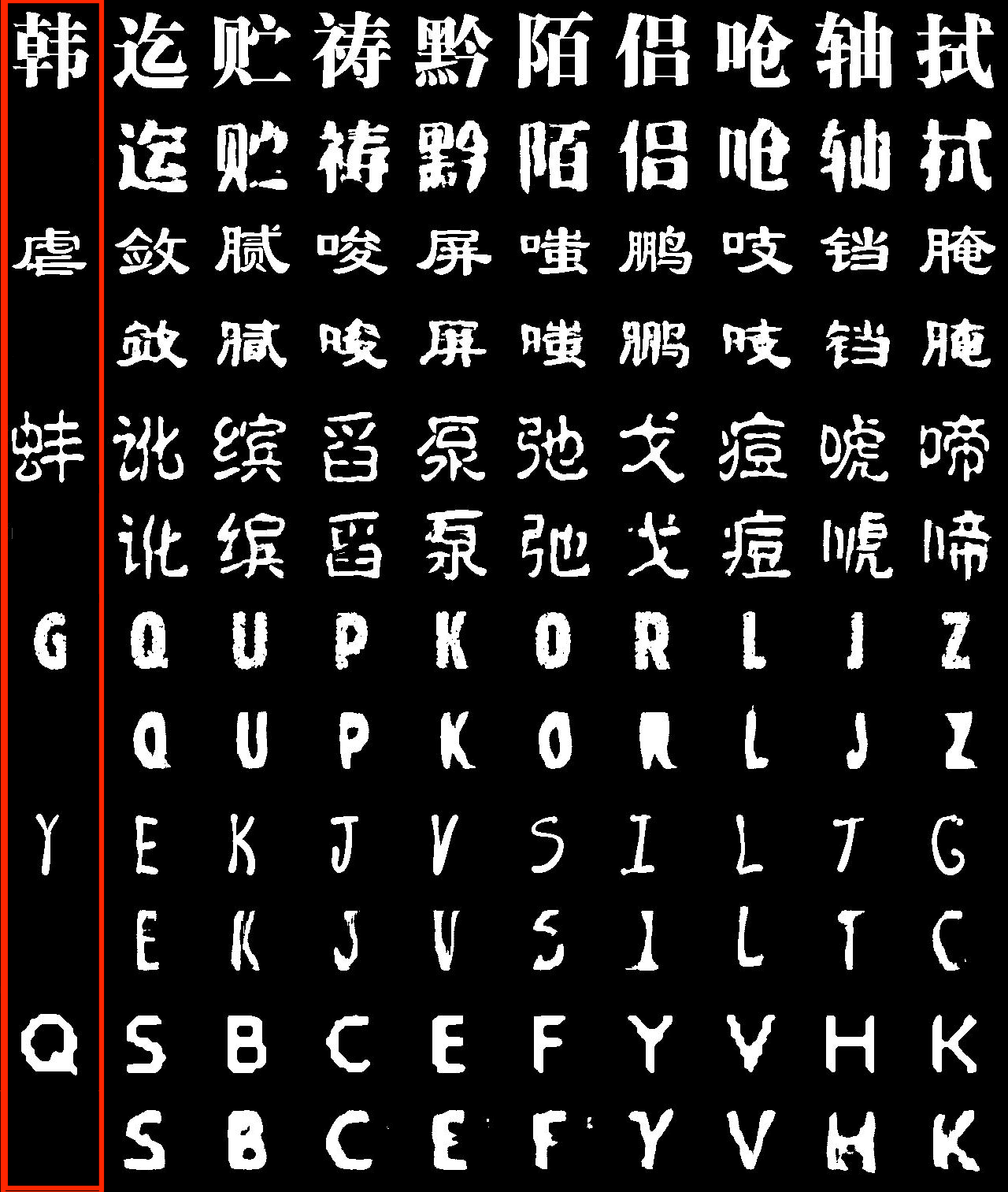}
  \caption{The figure consists of sections of which each two row represents the work on same typeface, respectively. Top left image is input image and second row are generated images in each section. The first row of each section are label images corresponding to second row.}

  \label{fig:tcn2}
\end{figure}

\begin{table*}[t]
\small
\begin{center}
\caption{Sample Pair Experimental Results}
\tabcolsep=0.19cm
\begin{tabular}[width=\columnwidth]{c|c|c|c|c|c|c|c|c|c|c|c|c}
\hline
\multirow{3}{*}{ } & \multicolumn{2}{c|}{\multirow{2}{*}{\#Parameter}} & \multicolumn{8}{c}{SSIM} \\\cline{4-13}
 & \multicolumn{2}{c|}{} & \multicolumn{4}{c|}{Typeface Completion} & \multicolumn{6}{c}{Reconstruction} \\ \cline{2-13}
 			& Ch	& Eng	& 598-370 & 598-268	& G-Q 	& G-Y	& 598	& 370	& 268 & G	& Q 	& Y 	 	\\\hline
$CycleGAN$ 	& \multicolumn{2}{c|}{2.8e+7 x $N^2$} 	& 0.5616& 0.5442 & 0.6915 & 0.5295 & 0.8740 & 0.8467 & 0.8670 & 0.9108 & 0.8941 & 0.8147  \\ \hline
$MUNIT$ 	& \multicolumn{2}{c|}{4.6e+7 x $N^2$} 	& 0.5438& 0.5229 & 0.6268 & 0.7028 & \textbf{0.9941} & \textbf{0.9956} & \textbf{0.9937} & \textbf{0.9962} & \textbf{0.9936} & \textbf{0.9960} \\ \hline
$StarGAN$ 	& 6.4e+7& 5.3e+7& 0.5506 & 0.5443 & 0.6985& 0.7274 & 0.5413 & 0.5813 & 0.5624 & 0.7463& 0.7026& 0.7944  \\ \hline
\multirow{2}{*}{$TCN$} 		& 5.6e+7& 2.4e+7& \multirow{2}{*}{\textbf{0.6673}} & \multirow{2}{*}{\textbf{0.6573}} & \multirow{2}{*}{\textbf{0.7959}}& \multirow{2}{*}{\textbf{0.8264}} & \multirow{2}{*}{0.6609} & \multirow{2}{*}{0.7011} & \multirow{2}{*}{0.6862} & \multirow{2}{*}{0.8116} & \multirow{2}{*}{0.8015} & \multirow{2}{*}{0.8604} \\ 
& (7.9e+7)& (4.5e+7)& &&&&&&&&& \\ \hline
\end{tabular}
\end{center}
\label{Sample_Experimental_Results}
\end{table*}

\begin{table*}[t]
\small
\begin{center}
\caption{Total Experimental Results}
\tabcolsep=0.15cm
\begin{tabular}[width=\columnwidth]{c|c|c|c|c|c|c|c|c|c|c|c|c|c|c|c|c}
\hline
\multirow{3}{*}{ } & \multicolumn{4}{c|}{SSIM} & \multicolumn{4}{c|}{L1} & \multicolumn{4}{c|}{Style Accuracy} & \multicolumn{4}{c}{Content Accuracy} \\\cline{2-17}
 & \multicolumn{2}{c|}{TC} & \multicolumn{2}{c|}{Reconst} & \multicolumn{2}{c|}{TC} & \multicolumn{2}{c|}{Reconst} & \multicolumn{2}{c|}{TC} & \multicolumn{2}{c|}{Reconst} & \multicolumn{2}{c|}{TC} & \multicolumn{2}{c}{Reconst}\\ \cline{2-17}
 & Ch & Eng & Ch & Eng & Ch & Eng & Ch & Eng & Ch & Eng & Ch & Eng & Ch & Eng & Ch & Eng\\\hline
$StarGAN$ 	& 0.568& 0.749& 0.575& 0.857& 0.212& 0.094& 0.191& 0.046& 0.009& 0.737& 0.009& 0.851& 0.402& \textbf{0.820}& 0.523& 0.951\\ \hline
$TCN$ 		& \textbf{0.653}& \textbf{0.794}& \textbf{0.676}& \textbf{0.949}& \textbf{0.163}& \textbf{0.088}& \textbf{0.060}& \textbf{0.014}& \textbf{0.645}& \textbf{0.891}& \textbf{0.802}& \textbf{0.998}& \textbf{0.531}& 0.819& \textbf{0.972}& \textbf{0.991} \\ \hline
\end{tabular}
\end{center}
\label{Experimental_Results}
\end{table*}

\section{Analysis}

 The reconstruction performance and the typeface completion performance of single-domain image-to-image models (CycleGAN, MUNIT) vary (Table 3) due to insufficient information of the features. When extracting features from character images, style and content features are duplicated or lost, not being disjoint, which is demonstrated by the translation results of these models. The output of the single-domain models is dependent on the input image, so the models achieve high performance in the reconstruction task where the target is the input. On the other hand, in the typeface completion task, the result appears to be a simple combination of inputs rather than a image-to-image translation. 

StarGAN obtained good performance on the general images of the image-to-image translation task, but not on the character dataset.
For the StarGAN, content accuracy is similar to our model, but typeface accuracy is not.
Because StarGAN uses the reconstruction loss to maintain typeface information, but it places a greater weight on content translation. The resulting image is fairly clean at the character level, but there is a limit to maintaining the typeface information of the input. (Figure \ref{fig:results}(c)).

%
%
%
%
Another difference is that, when calculating a loss in the one-hot vector, the cosine similarity of each vector is either one or zero.
Therefore, we can only determine whether two values are matched. 
To address this issue, we use a latent vector as a domain label which has continuous values for similarity scores between vectors. 
From the values, the vector determines the similarity and difference of the two vectors, which can help the classifier to learn.

Another difference between TCN and the other models is the use of input labels. Adding input labels for the model results in output images more similar to the real images.
%
%
Due to the differences described above, our model outperformed StarGAN by 10\% on typeface completion and 12\% on reconstruction at SSIM index, as shown in Table 5. And as shown in Figure \ref{fig:tcn2}, generated images have consistent typeface of input image. Even the results are unseen typefaces in the training process. Nonetheless, TCN generates an image similar to the target.

We also combined the one-hot label with the encoded latent vector rather than the original image. This improves the parameter efficiency of the model. In Table 4, the number of parameters of the TCN is the smallest. The number in parentheses include the parameters of the classifier used in the pre-train. This module is not used for main-train and inference, hence it is indicated separately. The number of parameters, except for this, is 12.5\% and 54\% decrease in Chinese and English, respectively, compared with StarGAN.

In ablation study, The full model showed the best performance quantitatively and qualitatively. SSIM loss had the greatest influence on SSIM index in the test set. However, the absence of SSIM loss has little affected another index beyond the SSIM index.
The elimination of identity loss did not affect typeface completion performance, but it was the most influential in the reconstruction task. 
These two losses are available because the dataset is paired. In the model, which eliminated both SSIM loss and identity loss, which did not utilize the target image, we obtained the intermediate result of the previous two studies. Because the two losses are complementary, the overall result is worse when there is one loss only.
The reconstruction loss and perceptual reconstruction loss associated with cycle consistency had similar effects on performance. In the absence of cycle consistency, we saw that typeface accuracy is increasing and content accuracy is significantly lowered.
Lastly, the absence of input content labels affected overall performance downgrades. As we can show from the qualitative comparison, it is involved in the detailed result without any significant difference from the result of the full model.

%
%
%
%

\begin{figure}[t]
  \centering
  \includegraphics[width=\columnwidth]{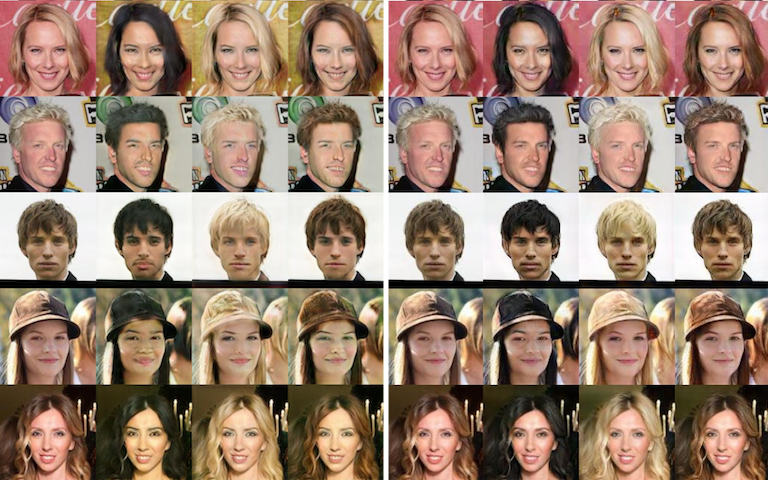}
  \caption{The results of the style transfer for CelebA. The four columns on the left and the right are the results of StarGAN and TCN.
  The first column of each result is the input, and the remaining columns are the result of changing the domain label to black / blond / brown hair, respectively.}
%
%
%
  \label{fig:genimg}
\end{figure}

\begin{figure}[t]
  \centering
  \includegraphics[width=\columnwidth]{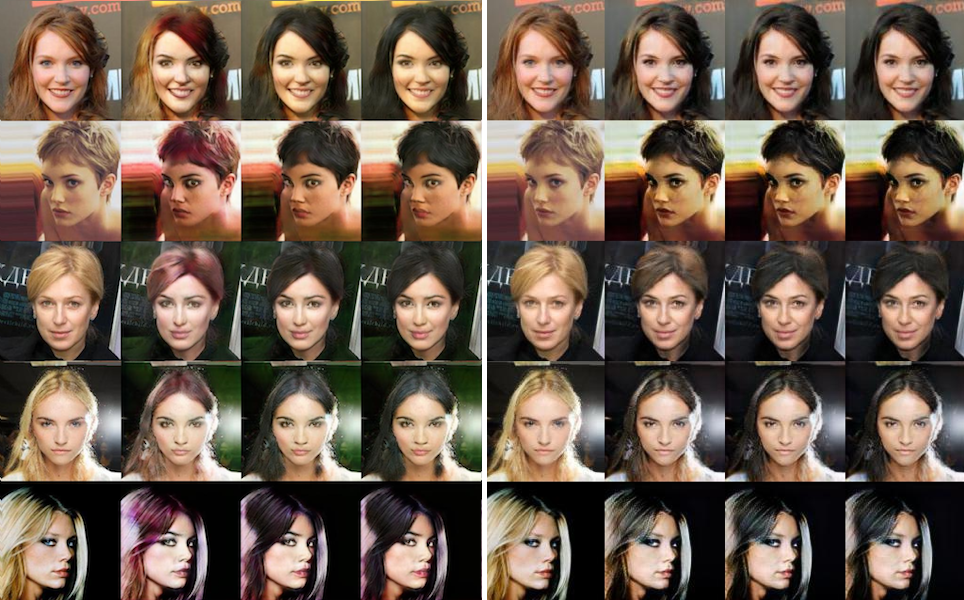}
  \caption{The result of the weighted style transfer from blond to black hair. The four columns on the left and right are the results of StarGAN and TCN, respectively. First column is the original image, and the weight of black color is assigned to 0.3, 0.6, and 0.9 for the rest. The hair color is somewhat reddish in StarGAN when the black hair weight is low, but TCN expresses the color between the blond and black hair based on the weight.}
%
%
%
  \label{fig:weighted_genimg}
\end{figure}

Our model can also be applied to unpaired general images. Since we do not have content information, We experimented after removing the content encoder. As shown in Figure \ref{fig:genimg}, we made a fairly plausible outcome. And when we compared the StarGAN, we found that our model maintains better out-of-style information that has not changed. We also were able to generate the image to maintain a specific ratio between the input and the target, like Figure \ref{fig:weighted_genimg}. This suggests that our model can be used for various applications.

\section{Conclusion}

In this paper, we proposed Typeface Completion Network (TCN) which generates an entire set of characters given only one characters while maintaining the typeface of the input characters. TCN utilizes the typeface and content encoders to effectively leverage the information of numerous classes. As a result, TCN learns multi-domain image-to-image translation using a single model, and produces more accurate outputs than existing baseline models. As illustrated in the qualitative analysis, we found that TCN successfully completes the character sets, which could reduce the costs of designing a new typeface. We also tested TCN on the CelebA dataset to demonstrate its applicability. In future work, we are planning a model that takes character subset as input and completes character set more finely.

\bibliography{sections/7_reference}
\bibliographystyle{aaai}

\end{document}